\documentclass{article}

\usepackage[utf8]{inputenc}
\usepackage{hyperref}
\usepackage{amsmath}
\usepackage{amssymb}
\usepackage{xcolor}
\usepackage{graphicx}
\usepackage{wrapfig}
\usepackage{hhline}
\usepackage{placeins}
\usepackage{multirow}

\usepackage[accepted]{icml2024}

\let\bld\boldsymbol

\icmltitlerunning{Boundary to Domain Mapping using LP-FNO for PDEs}
\begin{document}

\twocolumn[
\icmltitle{Learning the boundary-to-domain mapping using Lifting Product Fourier Neural Operators for partial differential equations}



\icmlsetsymbol{equal}{*}

\begin{icmlauthorlist}
\icmlauthor{Aditya Kashi}{ornl}
\icmlauthor{Arka Daw}{ornl}
\icmlauthor{Muralikrishnan Gopalakrishnan Meena}{ornl}
\icmlauthor{Hao Lu}{ornl}
\end{icmlauthorlist}

\icmlaffiliation{ornl}{Oak Ridge National Laboratory, Oak Ridge, TN 37830, USA}

\icmlcorrespondingauthor{Aditya Kashi}{kashia@ornl.gov}

\icmlkeywords{Neural operators, PDE, boundary value problem, lifting product, tensor product}

\vskip 0.3in
]



\printAffiliationsAndNotice{}

\begin{abstract}
Neural operators such as the Fourier Neural Operator (FNO) have been shown to provide resolution-independent deep learning models that can learn mappings between function spaces. For example, an initial condition can be mapped to the solution of a partial differential equation (PDE) at a future time-step using a neural operator. Despite the popularity of neural operators, their use to predict solution functions over a domain given only data over the boundary (such as a spatially varying Dirichlet boundary condition) remains unexplored. In this paper, we refer to such problems as boundary-to-domain problems; they have a wide range of applications in areas such as fluid mechanics, solid mechanics, heat transfer etc. We present a novel FNO-based architecture, named Lifting Product FNO (or LP-FNO) which can map arbitrary boundary functions defined on the lower-dimensional boundary to a solution in the entire domain. Specifically, two FNOs defined on the lower-dimensional boundary are lifted into the higher dimensional domain using our proposed lifting product layer. We demonstrate the efficacy and resolution independence of the proposed LP-FNO for the 2D Poisson equation.
\end{abstract}

\section{Introduction}

Computer simulations for real-world science and engineering problems can be complicated to set up, require knowledge of numerical methods, and may take a large amount of computational resources to compute a solution. Furthermore, there are some disciplines where the governing partial differential equations (PDEs) are not known with adequate certainty.
In these scenarios, surrogate models are attractive. These may be derived (simplified) from the governing equations by analysis or generated from data gathered in a laboratory or in the real world. In this work, we focus on data-driven surrogate models for phenomena governed by PDEs, particularly neural network-based methods. Neural networks that represent maps from the spatio-temporal domain (the region of interest) to the solution for one instance of a PDE, or family of PDEs parameterized by a few parameters, have been developed over the last several years. This has mainly been in the context of physics-informed neural networks (PINNs) \cite{raissi_physics-informed_2019, chen_physics-informed_nanooptics_2020} for both forward and inverse problems.
However, PINNs typically do not generalize to an adequately large parameter range \cite{krishnapriyan_pinn_fail_modes}. Further the initial and boundary conditions are baked into the model during training and cannot generalize when these are changed.

A different approach is to use an operator approach to map the entire input (a function over the spatial domain or its boundary) to the solution of the PDE, another function over the domain. For example, non-homogeneous material properties $a:\Omega \rightarrow \mathbb{R}$ can be mapped to the solution $u:\Omega \rightarrow \mathbb{R}$ of the heat equation. If the domain is a rectangle and the input function is discretized on a regular Cartesian grid, this can be done by convolutional neural networks.
The `DeepONet' architecture \cite{deeponet_2021} is an operator learning method that uses fully connected or convolutional neural networks to process the input function in the `branch' sub-network of their architecture, which can then be evaluated at different spatio-temporal coordinates using a separate `trunk' sub-network.
These operator approaches can have a much better payoff for the cost of training the models because the same model can now accept different spatially-varying parameter functions and initial conditions as input and predict the corresponding solution, rather than having them baked-in during training.
However, an issue with such models is that they are not fully `resolution-agnostic'; that is, they are designed for a fixed discretization of the input and output functions. Although the DeepONets are considered to be resolution-agnostic with regard to the output function that can be evaluated at any one point at each inference, the input to the branch network is still constrained to a particular discretization.
Furthermore, although adjustments can be made to convolution-based architectures to accept inputs of any size, they are typically not `resolution-independent', in that the accuracy suffers significantly if tested on resolutions unseen during training \cite{li_neural_operator_2020}.

Neural operators \cite{li_neural_operator_2020, kovachki_neural_op_2022} are a class of resolution-independent neural network-based operator models for PDE problems. One of these architectures, the Fourier Neural Operator (FNO) \cite{li_fourier_op_2021} has gained popularity in such fields as weather modeling \cite{fourcastnet,spherical_fno}. However, neural operators have been primarily explored for domain-to-domain mappings where the input is a function over the entire domain (typically an initial condition or material properties), and the output is over the entire domain as well (typically the solution at a later time). Despite the popularity of neural operators, their use to predict solution functions over a domain given only data over the boundary (such as a spatially varying Dirichlet boundary condition) remains unexplored. In this paper, we refer to such a problem as a boundary-to-domain (B2D) problem. This type of problems has a wide spectrum of applications in areas such as fluid mechanics, solid mechanics, heat transfer etc. 

To address the challenge, we aim to develop a Fourier neural operator (FNO) based architecture for a boundary-to-domain problem in PDE simulation. In other words, we ask the question: how can we predict the solution over the domain, given a function over the boundary (a boundary condition) as input? To this end, we propose a novel FNO-based architecture, named \emph{Lifting Product-FNO} (or LP-FNO) which can map arbitrary boundary functions defined on lower-dimensional manifolds to the entire solution domain. In particular, we extract hidden representations from two different FNOs defined on the lower-dimensional boundary domains, which are then lifted into the domain using a novel lifting product layer. For a simple 1D boundary to 2D domain problem, this lifting product operation is simply an outer product. 
In this paper, we demonstrate the B2D problem for the two-dimensional (2D) Poisson equation with non-homogeneous spatially-varying Dirichlet boundary conditions. We demonstrate the efficacy and resolution-independence of our proposed LP-FNO on this problem.

\begin{figure*}[h]
    \centering
    \includegraphics[width=0.8\linewidth]{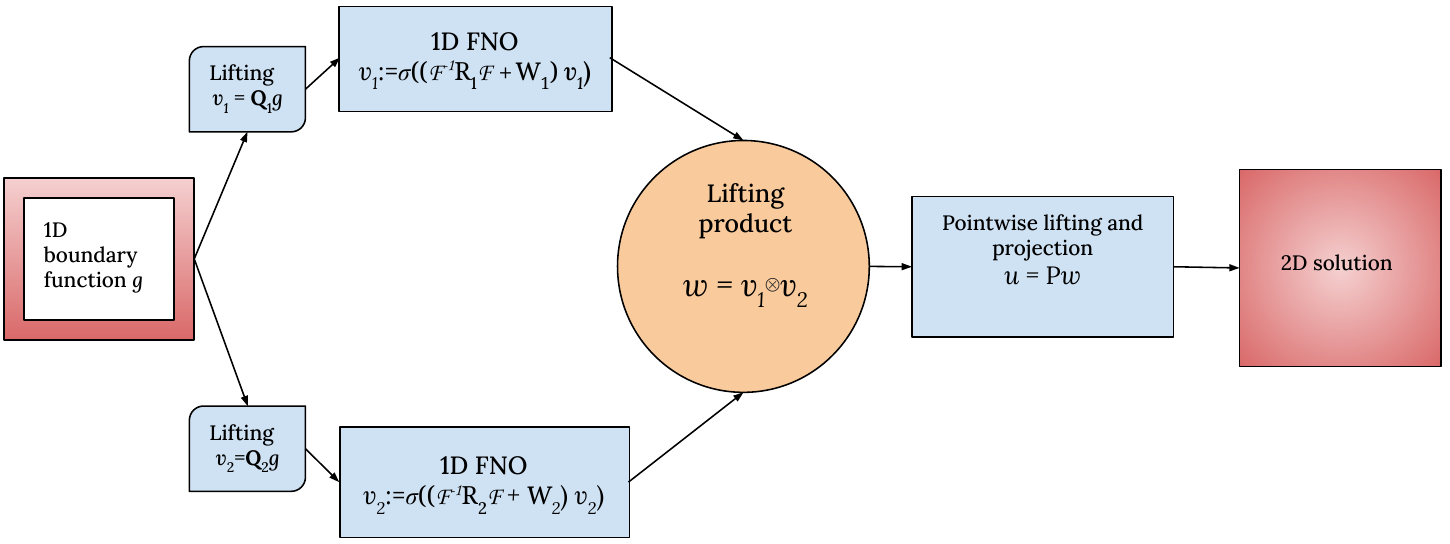}
    \caption{A schematic representation of the LP-FNO architecture example for the 1D to 2D case}
    \label{fig:lpfno}
\end{figure*}

\section{Boundary-to-Domain problem for PDEs}

We define the boundary to domain problem for PDEs in this section. Suppose $\Omega \subset \mathbb{R}^d$ represents a region in $d$-dimensional space (usually, $d=1,2,3$) with boundary $\partial\Omega$.
Let $\Gamma \subset L_2(\partial\Omega)$ be a sufficiently regular subset (e.g., having $d-1$ dimensional measure greater than zero) of the boundary. 
An arbitrary PDE can be represented as
\begin{equation}
    \bld{R}(\bld{u}(\bld{x},t)) = \bld{0}, \qquad \bld{x} \in \Omega, \, t \in [0,T],
    \label{eqn:pde}
\end{equation}
with boundary conditions
\begin{equation}
    \bld{b}(\bld{u}(\bld{x},t)) = \bld{0} \qquad \bld{x} \in \Gamma, \, t \in [0,T]
    \label{eqn:bcs}
\end{equation}
where $\bld{u} \in C([0,T], L^2(\Omega))^m$, the state, is a set of functions defined on the domain of interest ($\Omega \times [0,T]$) in space-time, $\bld{R}$ is a linear or nonlinear differential operator on the space of states, and $\bld{b}$ is an operator on the space of traces of the solution on $\Gamma$.
Steady-state scalar PDEs are those that do not depend on time and have a single variable of interest $u \in L^2(\Omega)$. An example is the Dirichlet problem for the Poisson equation:
\begin{align}
    -\nabla^2 u(\bld{x}) &= f, \qquad \bld{x} \in \Omega, \label{eqn:pde_poisson} \\
    u(\bld{x}) &= g \qquad \bld{x} \in \Gamma.
    \label{eqn:pde_poisson_bc}
\end{align}

In this paper, we define the boundary-to-domain problem for PDEs as the problem of predicting the $d$-dimensional solution function that obeys the PDE, given a ($d-1$)-dimension function over the boundary that represents the boundary conditions.
Our aim is to learn the map
\begin{equation}
    \mathcal{G}: \Gamma \rightarrow L^2(\Omega).
\end{equation}
In the case of the Poisson problem, this would map the boundary function $g$ in equation \eqref{eqn:pde_poisson_bc} to the solution $u$ over the domain.

\section{Lifting Product Fourier Neural Operator}

Let $\bld{g}:\Gamma \rightarrow \mathbb{R}^m$ be the input boundary function for $m$ physical variables (where $\Gamma \subset \partial\Omega$).
The schematic of our proposed Lifting-Product FNO (LP-FNO) is illustrated in figure \ref{fig:lpfno}, where we generate two separate feature representations of the input function. These are generated by two FNO blocks, each of which takes the boundary function $\bld{g}$ as the input.

FNO involves a hidden embedding space of dimension $n_e$ \cite{li_fourier_op_2021}. Assuming $\bld{g}$ is discretized by $N$ points, initially, hidden representations $\bld{h} \in \mathbb{R}^{N\times n_e}$ are generated by a learned point-wise lifting operator $\bld{Q}:\mathbb{R}^m \rightarrow \mathbb{R}^{n_e}$, not to be confused with the lifting product described later in this section:
\begin{equation}
    \bld{h}_0 = \bld{Q}\bld{g}.
\end{equation}

An FNO block consists of several FNO layers. Each FNO layer $\bld{F}_{\theta_i}: \mathbb{R}^{N \times n_e} \rightarrow \mathbb{R}^{N \times n_e}$ parameterized using ${\theta_i}$ for $i \in \{1, 2\}$ is given by \cite{li_fourier_op_2021}
\begin{align}
    \bld{F}_{\theta_i}(\bld{h}) &= \sigma(\mathcal{F}^{-1}(\bld{R}_i(\mathcal{F}(\bld{h}))) + \bld{W}_i\bld{h}) \\
    \text{or}, \quad \bld{F}_{\theta_i} &= \sigma(\mathcal{F}^{-1}\bld{R}_i\mathcal{F} + \bld{W}_i)
\end{align}
where $\mathcal{F}$ is the lower-dimensional real Fourier transform. $\bld{R}$ is a learned linear operator that acts independently on each mode of the Fourier transform but couples the different `channels' of the $n_e$-dimensional embedding space.
$\bld{W}_i:\mathbb{R}^{n_e} \rightarrow \mathbb{R}^{n_e}$ are point-wise linear neural network layers and $\sigma$ applies a scalar nonlinear activation to each hidden channel at each grid point.
We observe that the operations responsible for globally coupling all points in the spatial grid are the forward and inverse Fourier transforms.

Let $\bld{P}$ be a point-wise linear projection down from the embedding space to the state space of the PDE.
Then LP-FNO, with $l$ layers of lower-dimensional FNO in each FNO block, can be written as
\begin{align}
\bld{v}_i &= \prod_{j = 1}^{l} (\bld{F}_{\theta_{ij}}) \bld{Q}_i\, \bld{g} \quad i = 1,2; \\
\bld{w} &= \bld{v}_1 \otimes \bld{v}_2, \label{eq:lifting} \\
\bld{u} &= \bld{P} \bld{w},
\end{align}
where $\prod$ denotes sequential composition of functions.
Note that the lifting product $\otimes:L^2(\mathbb{R}^d) \times L^2(\mathbb{R}^d) \rightarrow L^2(\mathbb{R}^{d+1})$, to be defined below in the discrete sense, acts separately on each embedding dimension.
Therefore, in each embedding dimension, the inputs $\bld{v}_1, \bld{v}_2$ of equation \eqref{eq:lifting} are 1D vectors while the output $\bld{w}$ is a 2D `image' in space. This is the operation responsible for `lifting' the 1D hidden representation to 2D.
$\bld{W}_i, \bld{Q}$ and $\bld{P}$ all act separately at each point in physical space using feedforward neural networks.
The projection operator $\bld{P}:\mathbb{R}^{n_e} \rightarrow \mathbb{R}^m$ projects back to the space of physical variables. 

\par \noindent \textbf{Lifting Product:} Let $a$ and $b$ be two (discretized) functions on the lower-dimensional boundary of a rectangular domain. The tensor operation that lifts the input to a higher dimensional function $c$ can be written as
\begin{equation}
    c_{ij} = a_ib_j \quad \text{in 1D to 2D lifting}
\end{equation}
and
\begin{equation}
    c_{kij} = a_{ij}b_{ik} \quad \text{in 2D to 3D lifting}
\end{equation}
(no summation implied). The 1D to 2D operation is equivalent to an outer product $\bld{c} = \bld{a}\bld{b}^T$ separately on each channel. For the purpose of this paper, through some abuse of notation, we will denote both lifting product operations by $\bld{c} = \bld{a} \otimes \bld{b}$.

\section{Experimental Setup}

\subsection{Baselines}
\par \noindent \textbf{Resolution-agnostic Tencoder:}
As a baseline, we use a modification of Tencoder \cite{tencoder}, which is a convolutional neural network-based encoder-decoder architecture with a tensor product layer in the decoder. While the architecture was designed to train and test only on one fixed resolution, we modify it to work with varying input and output resolutions. Our modifications include using an adaptive average pooling layer in the encoder to downsample to a fixed-size latent space independent of the input size, as well as the use of bilinear interpolation in the decoder which enables upsampling to the required target output size. This enables the model trained on an arbitrary resolution to predict solutions for boundary functions sampled on uniform grids of different sizes, not just the one(s) it was trained on.

\par \noindent \textbf{FNO with zero padding:} We use a FNO-2d architecture \cite{li_fourier_op_2021}. However, that architecture was designed for a domain-to-domain scenario and requires a function over the entire domain as input. To address this, the input to the FNO is the 1D-boundary function $g$ padded with zeros all over the 2D-domain $\Omega$.

\begin{table*}[ht]
    \centering
\begin{tabular}{|l|l|ll|llllll|}
\hline
\multirow{2}{*}{\textbf{}}    & \multirow{2}{*}{\textbf{}} & \multicolumn{2}{c|}{\multirow{2}{*}{\textbf{In-Distribution}}} & \multicolumn{6}{c|}{\textbf{Out-of-Distribution}}                                                                                                                                                                     \\ \cline{5-10} 
                              &                            & \multicolumn{2}{l|}{}                                          & \multicolumn{2}{c|}{\textbf{Gaussian}}                                    & \multicolumn{2}{c|}{\textbf{Sinusoidal}}                                        & \multicolumn{2}{c|}{\textbf{Polynomial}}                \\ \hline
\textbf{Resolution}           & \textbf{Model}             & \multicolumn{1}{l|}{\textbf{Rel. L1}}     & \textbf{Rel L2}    & \multicolumn{1}{l|}{\textbf{Rel. L1}} & \multicolumn{1}{l|}{\textbf{Rel L2}} & \multicolumn{1}{l|}{\textbf{Rel. L1}} & \multicolumn{1}{l|}{\textbf{Rel L2}} & \multicolumn{1}{l|}{\textbf{Rel. L1}} & \textbf{Rel L2} \\ \hline
\multirow{3}{*}{\textbf{32}}  & Tencoder          & \multicolumn{1}{l|}{0.01257}              & 0.00426            & \multicolumn{1}{l|}{0.90621}          & \multicolumn{1}{l|}{0.40405}         & \multicolumn{1}{l|}{1.74984}          & \multicolumn{1}{l|}{0.94939}         & \multicolumn{1}{l|}{0.91356}          & 0.44197         \\ \cline{2-10} 
                              & FNO2d             & \multicolumn{1}{l|}{0.00254}              & \textbf{0.00139}            & \multicolumn{1}{l|}{0.73337}          & \multicolumn{1}{l|}{0.39078}         & \multicolumn{1}{l|}{1.11024}          & \multicolumn{1}{l|}{0.73673}         & \multicolumn{1}{l|}{0.71}             & 0.3975          \\ \cline{2-10} 
                              & TP-FNO            & \multicolumn{1}{l|}{0.0108}               & 0.00463            & \multicolumn{1}{l|}{0.30713}          & \multicolumn{1}{l|}{\textbf{0.05389}}         & \multicolumn{1}{l|}{0.52667}          & \multicolumn{1}{l|}{\textbf{0.15851}}         & \multicolumn{1}{l|}{0.3259}           & \textbf{0.06815}         \\ \hline
\multirow{3}{*}{\textbf{64}}  & Tencoder          & \multicolumn{1}{l|}{0.04765}              & 0.0238             & \multicolumn{1}{l|}{0.03826}          & \multicolumn{1}{l|}{0.02354}         & \multicolumn{1}{l|}{0.23802}          & \multicolumn{1}{l|}{0.15035}         & \multicolumn{1}{l|}{0.05085}          & 0.02518         \\ \cline{2-10} 
                              & FNO2d             & \multicolumn{1}{l|}{0.00541}              & \textbf{0.0019}             & \multicolumn{1}{l|}{0.00347}          & \multicolumn{1}{l|}{\textbf{0.00126}}         & \multicolumn{1}{l|}{0.06382}          & \multicolumn{1}{l|}{\textbf{0.0191}}          & \multicolumn{1}{l|}{0.0142}           & \textbf{0.00435}         \\ \cline{2-10} 
                              & TP-FNO            & \multicolumn{1}{l|}{0.01937}              & 0.0064             & \multicolumn{1}{l|}{0.00542}          & \multicolumn{1}{l|}{0.0016}          & \multicolumn{1}{l|}{0.19085}          & \multicolumn{1}{l|}{0.0889}          & \multicolumn{1}{l|}{0.04098}          & 0.01268         \\ \hline
\multirow{3}{*}{\textbf{128}} & Tencoder                   & \multicolumn{1}{l|}{0.15391}              & 0.0682             & \multicolumn{1}{l|}{1.14888}          & \multicolumn{1}{l|}{0.53499}         & \multicolumn{1}{l|}{2.21484}          & \multicolumn{1}{l|}{1.2091}          & \multicolumn{1}{l|}{1.15768}          & 0.56814         \\ \cline{2-10} 
                              & FNO2d                      & \multicolumn{1}{l|}{0.01125}              & \textbf{0.00271}            & \multicolumn{1}{l|}{2.07062}          & \multicolumn{1}{l|}{1.376}           & \multicolumn{1}{l|}{2.15543}          & \multicolumn{1}{l|}{1.6864}          & \multicolumn{1}{l|}{1.96741}          & 1.3438          \\ \cline{2-10} 
                              & TP-FNO                     & \multicolumn{1}{l|}{0.03501}              & 0.0082             & \multicolumn{1}{l|}{0.05156}          & \multicolumn{1}{l|}{\textbf{0.01847}}         & \multicolumn{1}{l|}{0.3278}           & \multicolumn{1}{l|}{\textbf{0.13745}}         & \multicolumn{1}{l|}{0.111}            & \textbf{0.0374}          \\ \hline
\end{tabular}
    \caption{Relative L1 and L2 norm errors of predictions from four models trained on data at different resolutions. The in-distribution (I.D.) test data are at each model's `native' resolution (the one it was trained on) while the out-of-distribution (O.O.D.) test data are at a 64x64 resolution for all the models in this table.}
    \label{tab:r1}
\end{table*}

\begin{figure}[h]
    \includegraphics[width=1.05\linewidth]{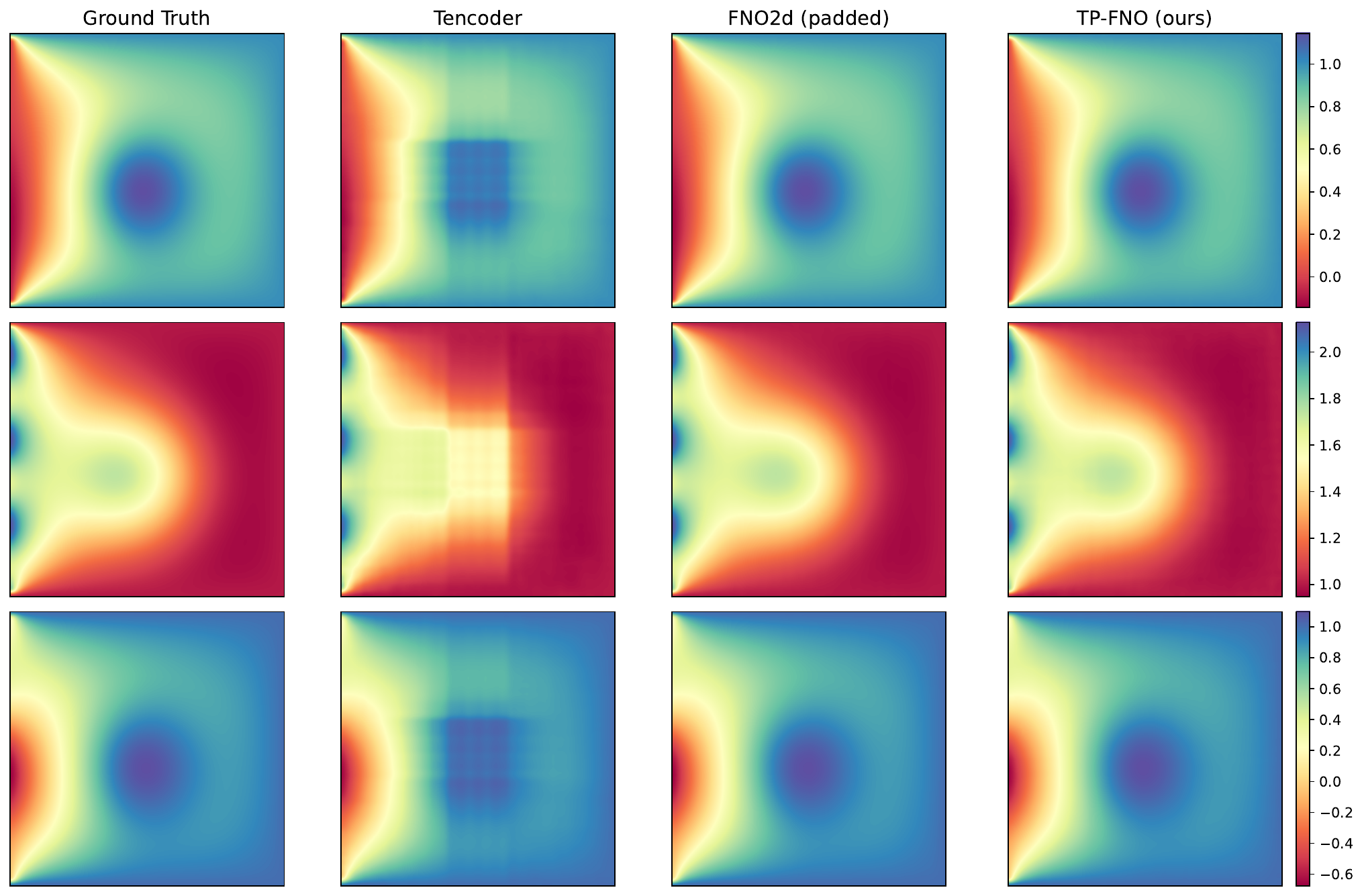}
    \caption{Comparison of the predicted solutions on in-distribution examples for models trained on 64x64 resolution.}
    \label{fig:field_in_dist_64}
\end{figure}

\begin{figure}[h]
    \includegraphics[width=1\linewidth]{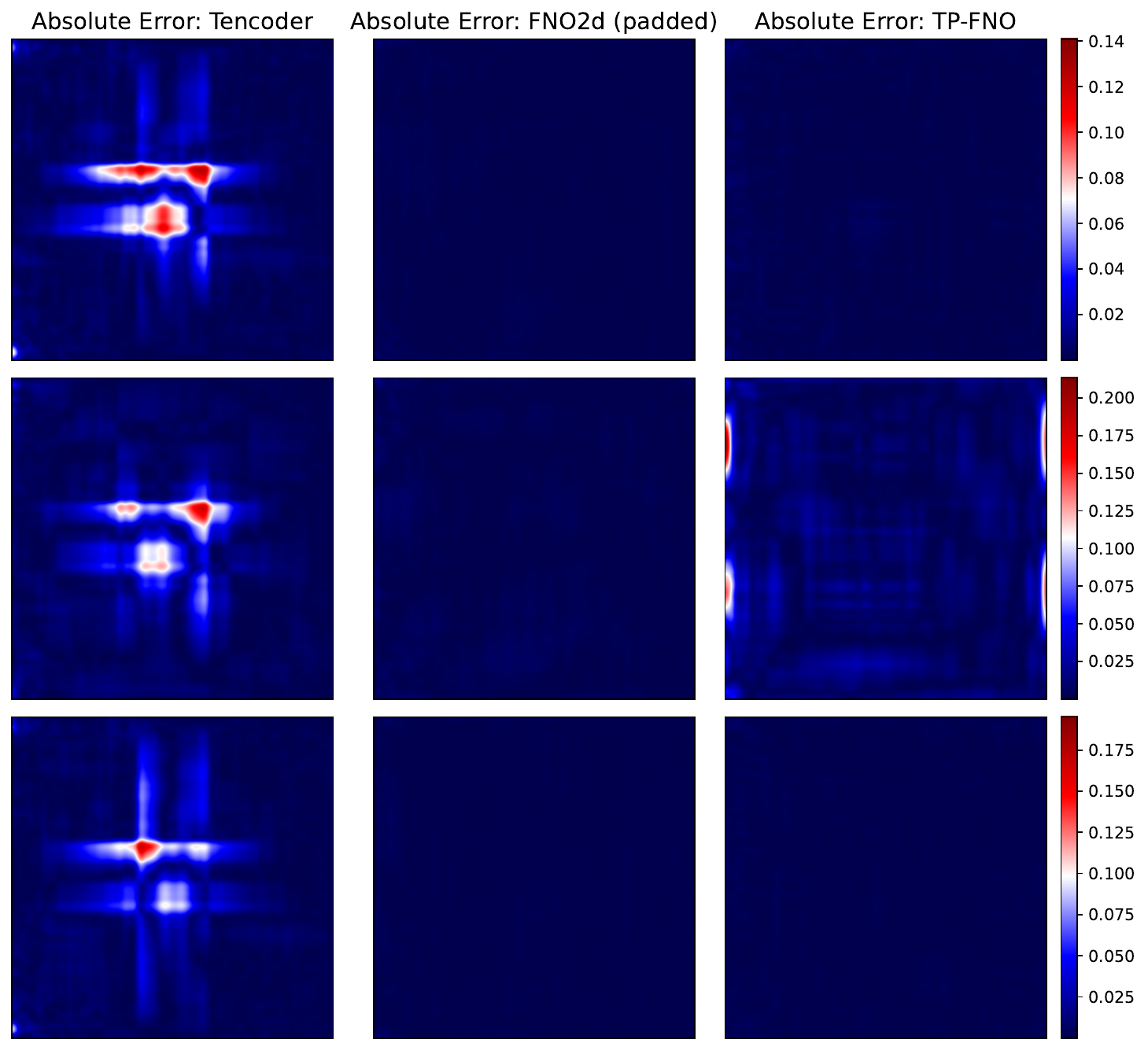}
    \caption{Comparison of the absolute error on in-distribution examples for models trained on 64x64 resolution.}
    \label{fig:mae_in_dist_64}
\end{figure}

\subsection{Problem setup}

We solve the boundary-to-domain problem for the Poisson equation \eqref{eqn:pde_poisson} with Dirichlet conditions (equation \eqref{eqn:pde_poisson_bc}). The problem setup is identical to the one used to evaluate Tencoder \cite{tencoder}.
In summary, three families of boundary functions are specified on the left boundary of a square domain and simulations run to obtain training and test datasets. 
The boundary functions are parameterized by a few parameters (this parameterization is never exposed to the model), and there are separate test sets containing in-distribution and out-of-distribution samples with respect to the training set.
The training set consists of 2048 samples (pairs of boundary condition and corresponding solution) for two families of boundary functions - Gaussian and sinusoidal. We test on in-distribution and out-of-distribution samples from those two families but also from a third family of polynomial functions up to degree 4.
We train the models with data on some fixed resolution, and test it on data on both the same resolution (``native resolution'') and also other resolutions it has not seen during training (``non-native resolutions'').

\begin{table}[]
\begin{tabular}{|l|l|lll|}
\hline
\multirow{2}{*}{\textbf{\begin{tabular}[c]{@{}l@{}}Training\\  Resolution\end{tabular}}} & \multirow{2}{*}{\textbf{Model}} & \multicolumn{3}{l|}{\textbf{Testing Resolution}}                                   \\ \cline{3-5} 
                &                                  & \multicolumn{1}{l|}{\textbf{32}} & \multicolumn{1}{l|}{\textbf{64}} & \textbf{128} \\ \hline
\multirow{3}{*}{\textbf{32}}  & Tencoder           & \multicolumn{1}{l|}{0.00426}     & \multicolumn{1}{l|}{0.34346}     & 0.3282       \\ \cline{2-5} 
                              & FNO2d      & \multicolumn{1}{l|}{\textbf{0.00139}}     & \multicolumn{1}{l|}{0.29973}     & 0.3535       \\ \cline{2-5} 
                              & TP-FNO     & \multicolumn{1}{l|}{0.00463}     & \multicolumn{1}{l|}{\textbf{0.05119}}     & \textbf{0.09289}      \\ \hline
\multirow{3}{*}{\textbf{64}}  & Tencoder   & \multicolumn{1}{l|}{0.41229}     & \multicolumn{1}{l|}{0.0238}      & 0.32779      \\ \cline{2-5} 
                              & FNO2d      & \multicolumn{1}{l|}{0.6804}      & \multicolumn{1}{l|}{\textbf{0.0019}}      & 0.58979      \\ \cline{2-5} 
                              & TP-FNO     & \multicolumn{1}{l|}{\textbf{0.05879}}     & \multicolumn{1}{l|}{0.0064}      & \textbf{0.04006}      \\ \hline
\multirow{3}{*}{\textbf{128}} & Tencoder   & \multicolumn{1}{l|}{0.46239}     & \multicolumn{1}{l|}{0.42954}     & 0.0682       \\ \cline{2-5} 
                              & FNO2d      & \multicolumn{1}{l|}{2.29772}     & \multicolumn{1}{l|}{1.17268}     & \textbf{0.00271}      \\ \cline{2-5} 
                              & TP-FNO     & \multicolumn{1}{l|}{\textbf{0.09778}}     & \multicolumn{1}{l|}{\textbf{0.02646}}     & 0.0082       \\ \hline
\end{tabular}
    \caption{Relative L2 norms of error on resolution-independence on in-distribution test sets.}
    \label{tab:res-indep}
\end{table}

\begin{figure} [ht]
    \includegraphics[width=1.05\linewidth]{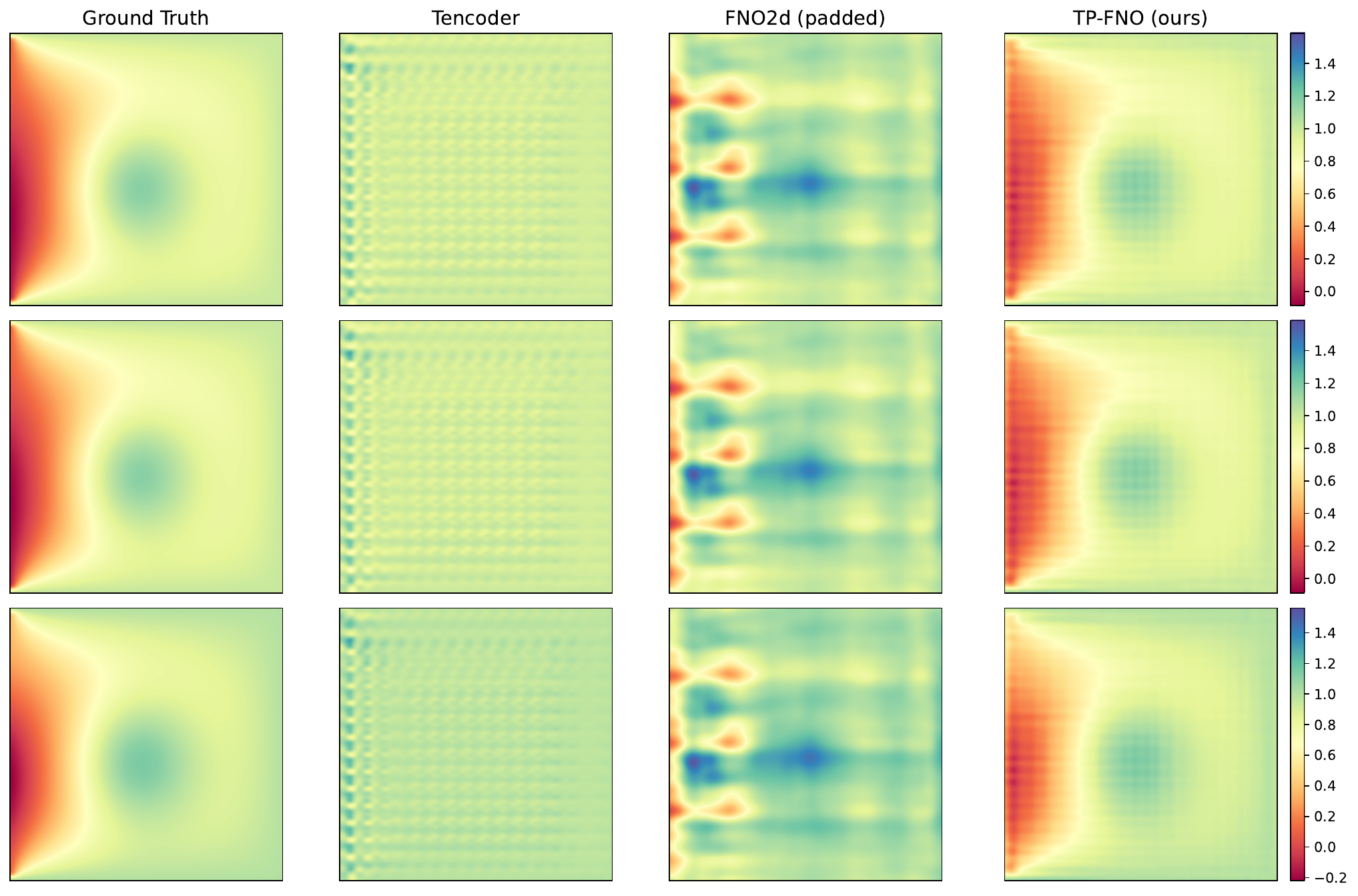}
    \caption{Comparison of the predicted solutions on out-of-distribution exponential examples for models trained on 32x32 resolution and evaluated on 64x64.}
    \label{fig:field_ood_dist_32}
\end{figure}

\begin{figure}[ht]
    \includegraphics[width=1\linewidth]{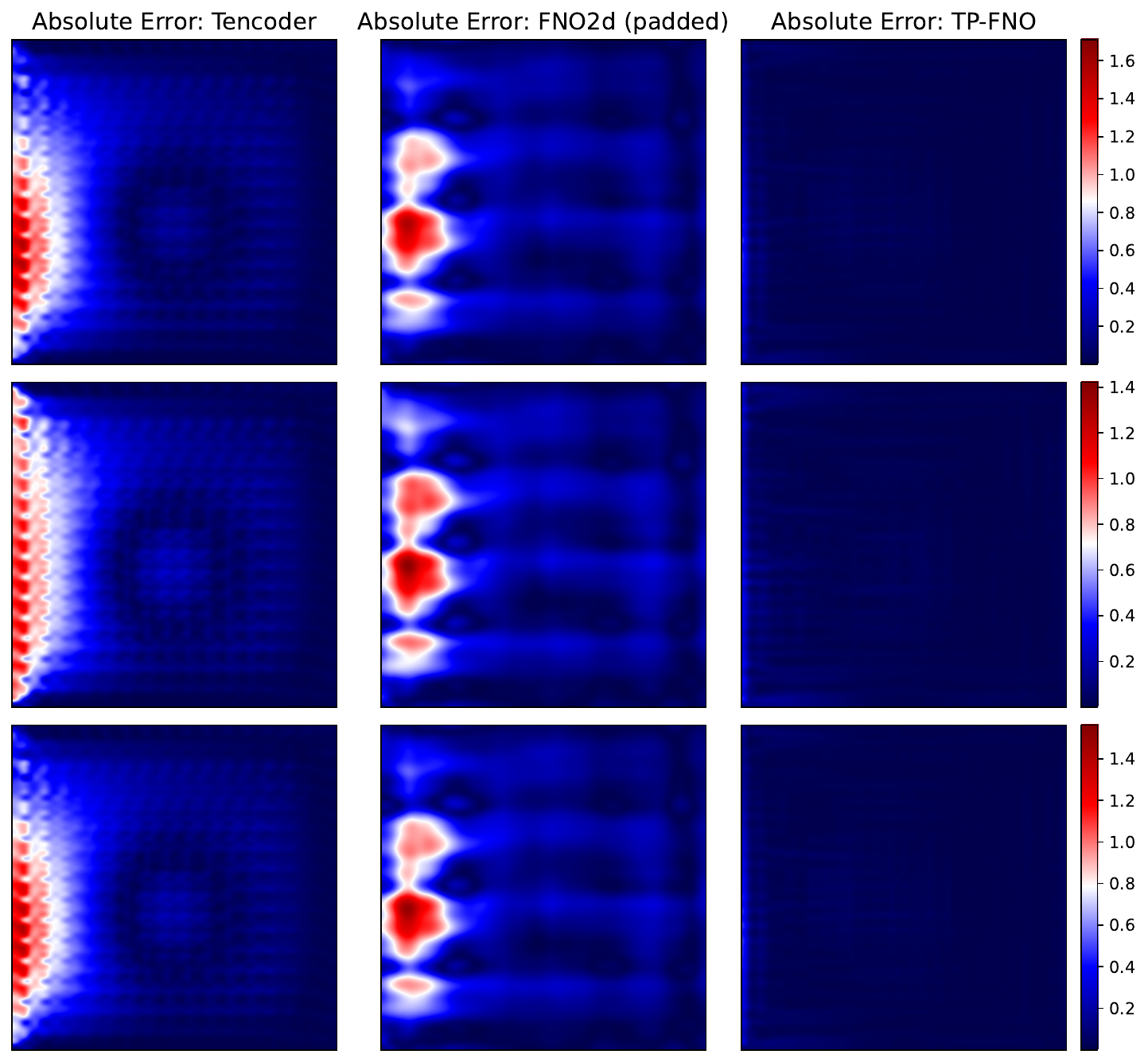}
    \caption{Comparison of the absolute error on out-of-distribution exponential examples for models trained on 32x32 resolution and evaluated on 64x64.}
    \label{fig:mae_ood_dist_32}
\end{figure}

\begin{figure}[ht]
    \includegraphics[width=1.05\linewidth]{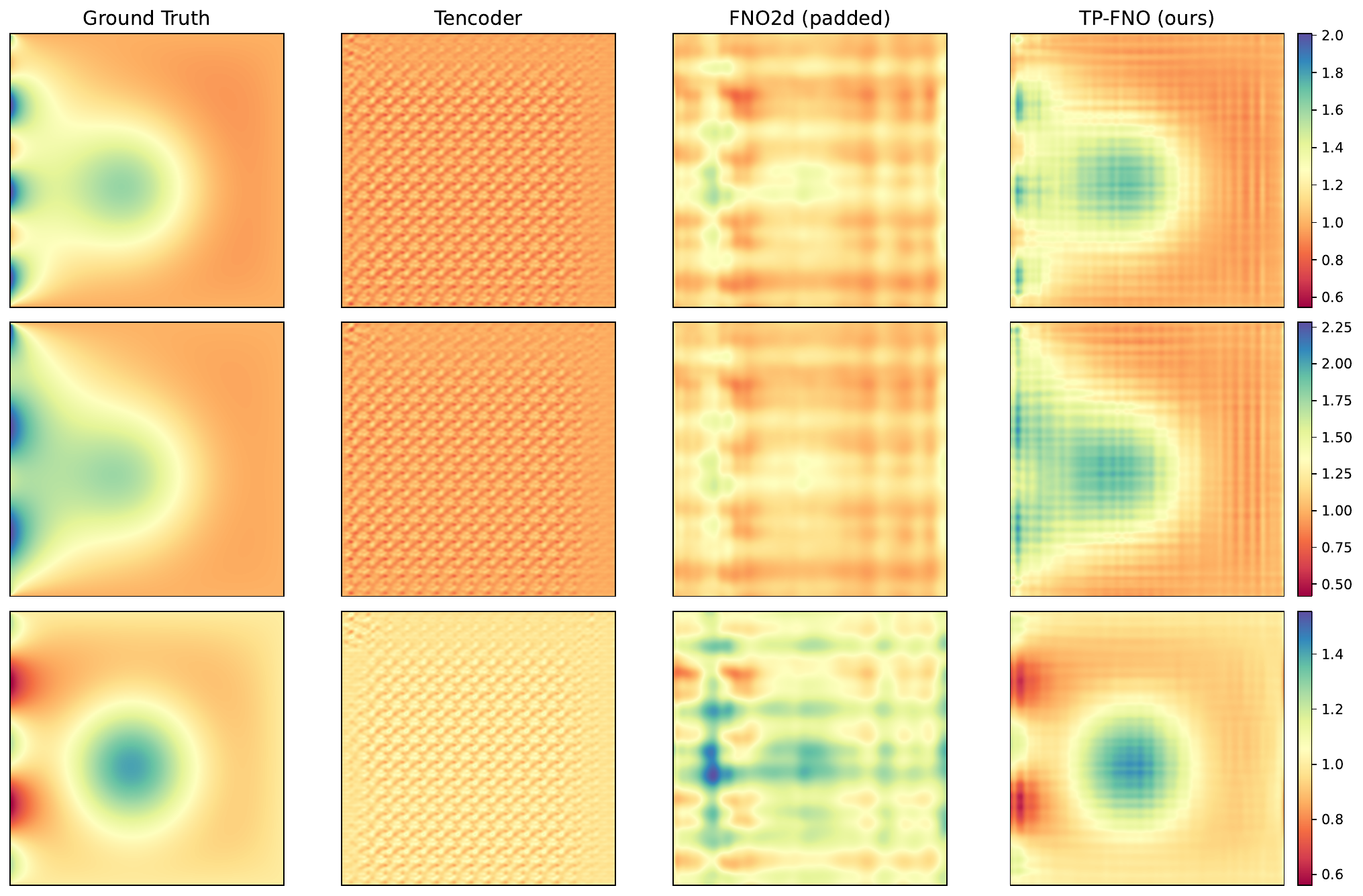}
    \caption{Comparison of the predicted solutions for models trained on 32x32 resolution and evaluated on 128x128.}
    \label{fig:superres_32_128}
\end{figure}

\section{Results}
In Table \ref{tab:r1}, we first show results from the three architectures on in-distribution samples on the same resolution they were trained on, and out-of-distribution (OOD) samples on a 64x64 grid. All the models were trained for 200 epochs over the training set comprising 2048 samples.

\par \noindent \textbf{In-distribution performance of LP-FNO with baselines:}
On the in-distribution test samples, we observe that the FNO2d (padded with zeros) and our proposed LP-FNO shows similar accuracy in around $10^{-3}$ relative L2 errors, with FNO-2d having a slight edge in performance for resolutions $64\times64$ and $128\times128$. The Tencoder consistently performs significantly worse that the other two models, with almost an order of magnitude difference in performance. Three example predictions of the solutions from the test set and their respective absolute errors are shown in Figure \ref{fig:field_in_dist_64} and Figure \ref{fig:mae_in_dist_64} respectively. Similar to our previous observation, we see that both the PDE solutions and the absolute errors for the FNO-2d and LP-FNO are comparable, while the predictions of Tencoder demonstrate checkerboard artefacts which are also visible in the absolute error plots. See more visualization of in-distribution examples in the Appendix.

\par \noindent \textbf{Out-of-Distribution Generalization of LP-FNO:} However, on out-of-distribution test sets on non-native resolution of 64x64 (the first and third row-blocks), padded FNO2d performs poorly, with relative errors greater than 1.
In comparison, LP-FNO still retains a fair accuracy, with eg., ~1.8\% accuracy in relative L2 error with a model that is trained on 128x128 data and inferenced on 64x64 Gaussian samples. In that example, padded 2D FNO has a huge error rate of 137\%. While this result for the FNO2d needs to be investigated further and corroborated, it points to a loss of resolution-independence due to the zero-padding required in input processing. As expected, the resolution-agnostic Tencoder, while being able to operate seamlessly on inputs of any resolution, does not perform well on non-native resolutions.

Figure \ref{fig:field_ood_dist_32} and \ref{fig:mae_ood_dist_32} present the out-of-distribution solution fields and absolute errors of models trained on $32\times32$ and inferenced on out-of-distribution $64$-dimensions inputs from the Gaussian family. We observe that the predictions for both the Tencoder and the FNO2d are extremely poor. While, the solutions obtained by of our proposed LP-FNO has some artefacts, it was able to capture some of the high-level patterns present in the solution. These high-level similarities in the solution demonstrate the resolution-independence capabilities of the LP-FNO, on out-of-distribution sample, which is a challenging task. Further, we believe that these artefacts shown by LP-FNO can be improved with better hyperparameter tuning and modifications to the LP-FNO model architecture as detailed in section \ref{sec:concl}. See more visualization of out-of-distribution examples for sinusoidal and polynomial cases in the appendix.

\par \noindent \textbf{Resolution Independence of LP-FNO:} Table \ref{tab:res-indep} shows the degree to which each of the architectures is resolution-independent using test sets which are in-distribution in terms of boundary function parameters. Here, again, we see that on native resolutions (diagonal blocks in the table), all of the three models are able to learn good solutions, with FNO2d and TP-FNO being approximately an order of magnitude better than Tencoder. However, when tested on non-native resolution, the resolution independence fails to hold for Tencoder and FNO2d and we observe very large L2 errors. In contrast, LP-FNO demonstrates a reasonable accuracy even on non-native resolutions.
We also visualize the zero-shot super-resolution capabilities of the three models in Figure \ref{fig:superres_32_128}, where the models are trained on $32\times32$ and then evaluated on $128\times128$. As expected, we observe that the Tencoder and FNO2d fails to perform zero-shot super-resolution, while our proposed LP-FNO is able to capture the high-level details of the solution functions. However, it must be noted that LP-FNO still shows significant checkerboard artefacts in the solutions, and our future work would focus on that.

\section{Conclusion and Future Work}
\label{sec:concl}

We have presented LP-FNO, a neural operator based on FNO and a lifting product operation and demonstrated its potential on a simple 2D model problem. 
While our architecture still demonstrates some artefacts in its solution for out-of-domain and resolution independence tasks, it generalizes well to other resolutions, which puts it on the right track in terms of fulfilling the promise of neural operators for boundary-to-domain problems in PDE surrogate models.

The architecture presented is still a preliminary work part of ongoing developments. In the near future, several improvements are planned which are expected to significantly improve accuracy and efficiency on both native and non-native resolutions.
\begin{itemize}
\item Our aim is to implement LP-FNO with tensor product in modal (or frequency) space rather than physical space. We expect this to make the architecture more scalable while improving its accuracy.
\item We will investigate the effect of adding 2D FNO layers after the tensor product layer in addition to the simple linear projection layer used in the current implementation.
\item We will test the architecture on nonlinear PDEs and experiment with different activation functions.
\item We intend to prove theoretically that our architecture has the universal approximation property in a relevant function space and that it is resolution independent.
\end{itemize}

\section{Impact statement}
This paper presents work whose goal is to advance the field of machine learning. There are many potential societal consequences of our work, none which we feel must be specifically highlighted here. The eventual most direct impact will be in achieving transformational acceleration of computer simulations in science and engineering.

\section*{Acklowledgements}
This manuscript has been authored by UT-Battelle, LLC, under contract DE-AC05-00OR22725 with the US Department of Energy (DOE). The US government retains and the publisher, by accepting the article for publication, acknowledges that the US government retains a nonexclusive, paid-up, irrevocable, worldwide license to publish or reproduce the published form of this manuscript, or allow others to do so, for US government purposes. DOE will provide public access to these results of federally sponsored research in accordance with the DOE Public Access Plan (https://www.energy.gov/doe-public-access-plan).

\bibliography{refs}
\bibliographystyle{icml2024}

\onecolumn

\appendix
\twocolumn
\section{Additional results}
\FloatBarrier
\subsection{Visualization of Other In-Distribution Results}
\begin{figure}[h]
    \includegraphics[width=1\linewidth]{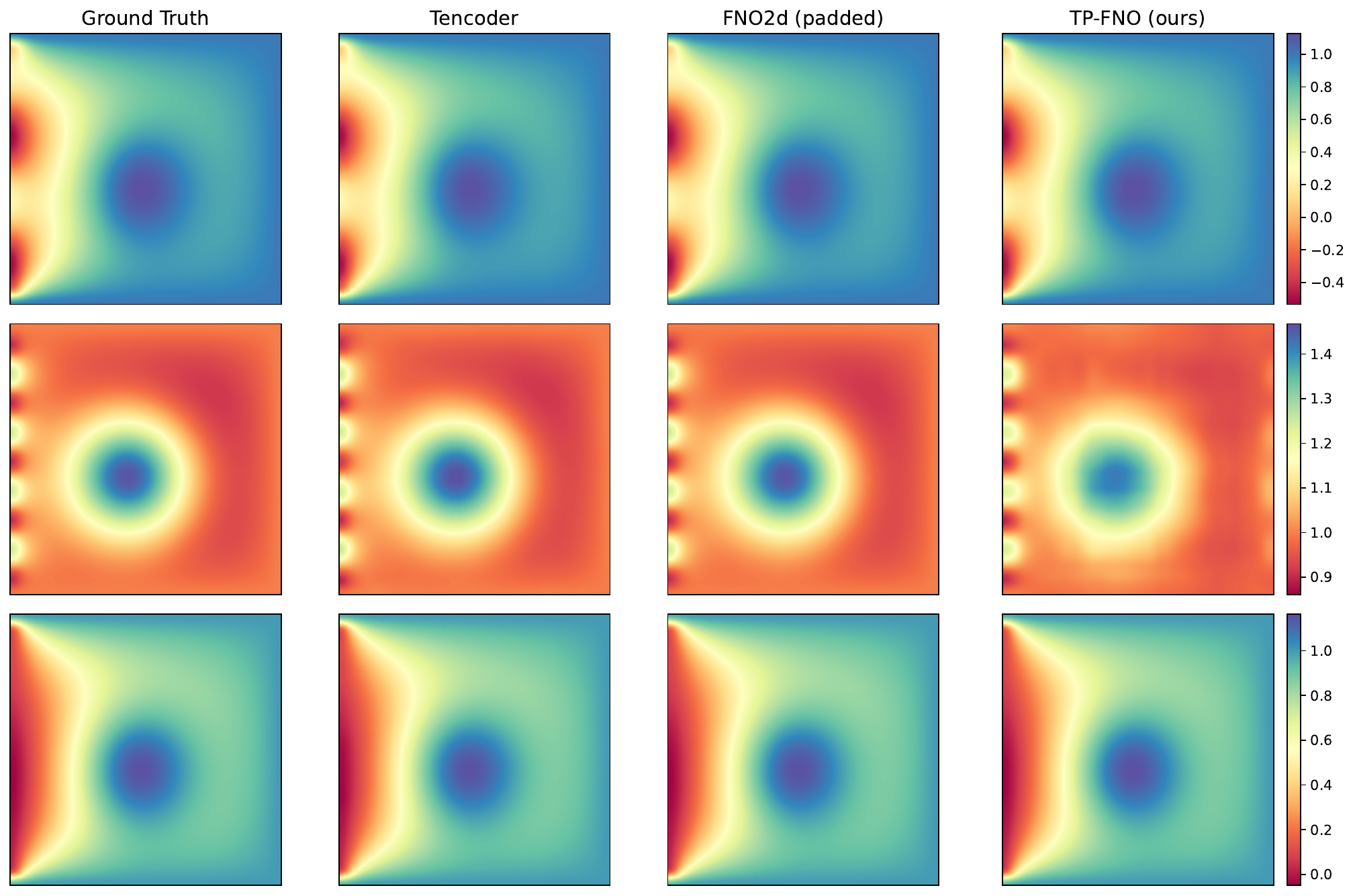}
    \caption{Comparison of the predicted solutions on in-distribution examples for models trained on 32x32 resolution.}
    \label{fig:field_in_dist_32}
\end{figure}

\begin{figure}[h]
    \includegraphics[width=0.9\linewidth]{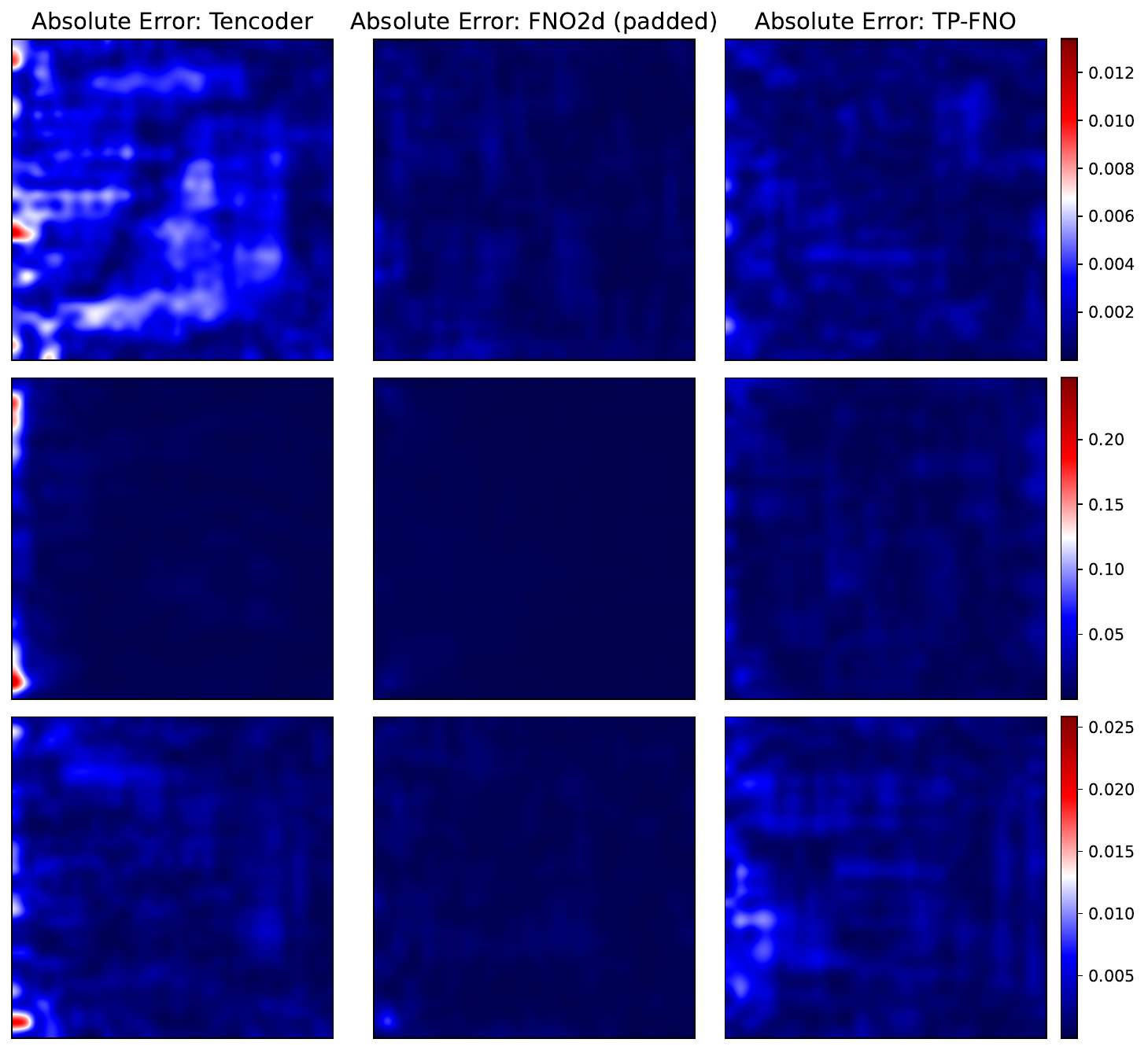}
    \caption{Comparison of the absolute error on in-distribution examples for models trained on 32x32 resolution.}
    \label{fig:mae_in_dist_32}
\end{figure}

Additional visualizations of the model performances on 32x32 can be seen in Figures \ref{fig:field_in_dist_32} and \ref{fig:mae_in_dist_32}.

\begin{figure}[h]
    \includegraphics[width=1\linewidth]{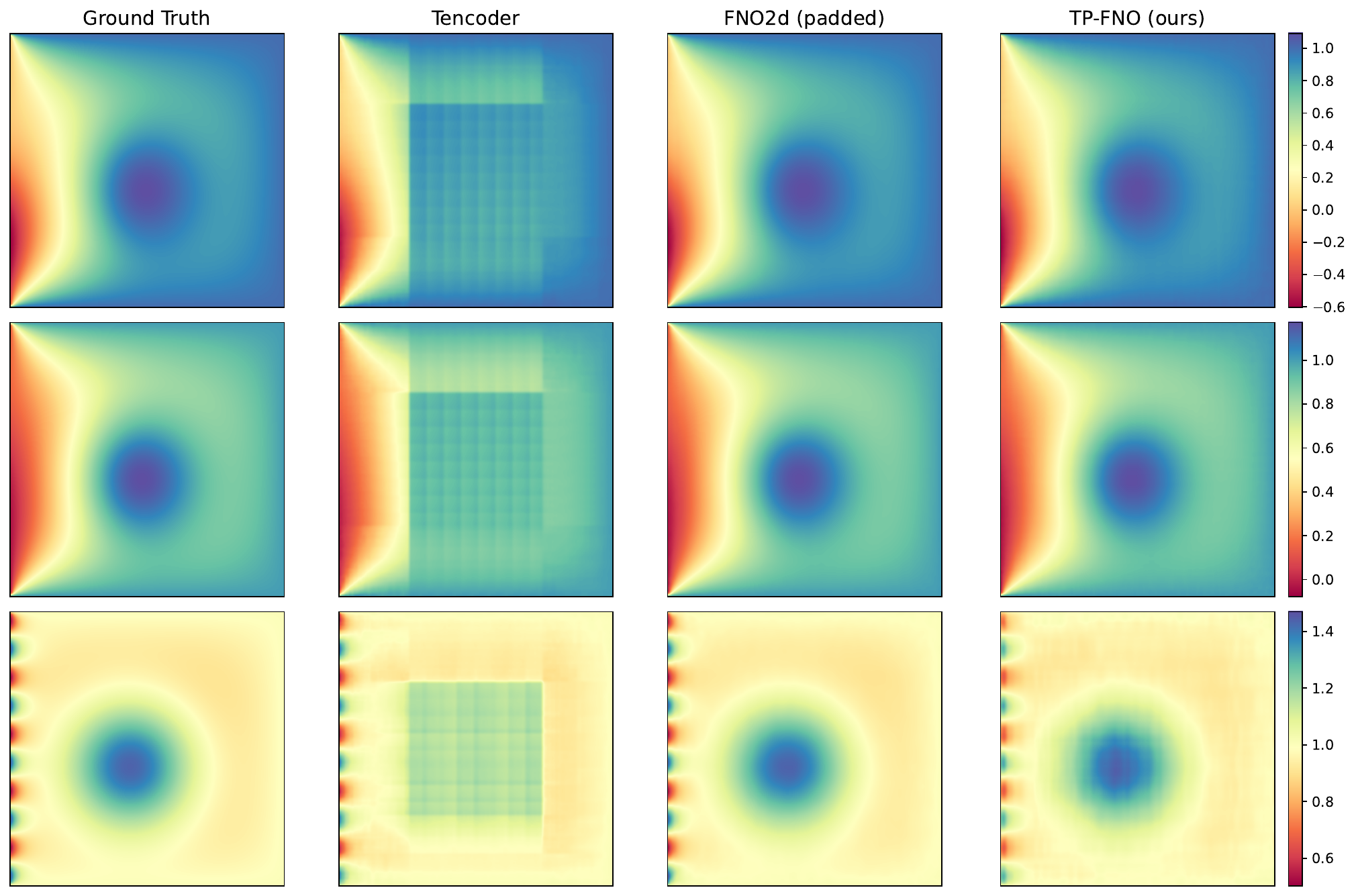}
    \caption{Comparison of the predicted solutions on in-distribution examples for models trained on 128x128 resolution.}
    \label{fig:field_in_dist_128}
\end{figure}

\begin{figure}[h]
    \includegraphics[width=0.92\linewidth]{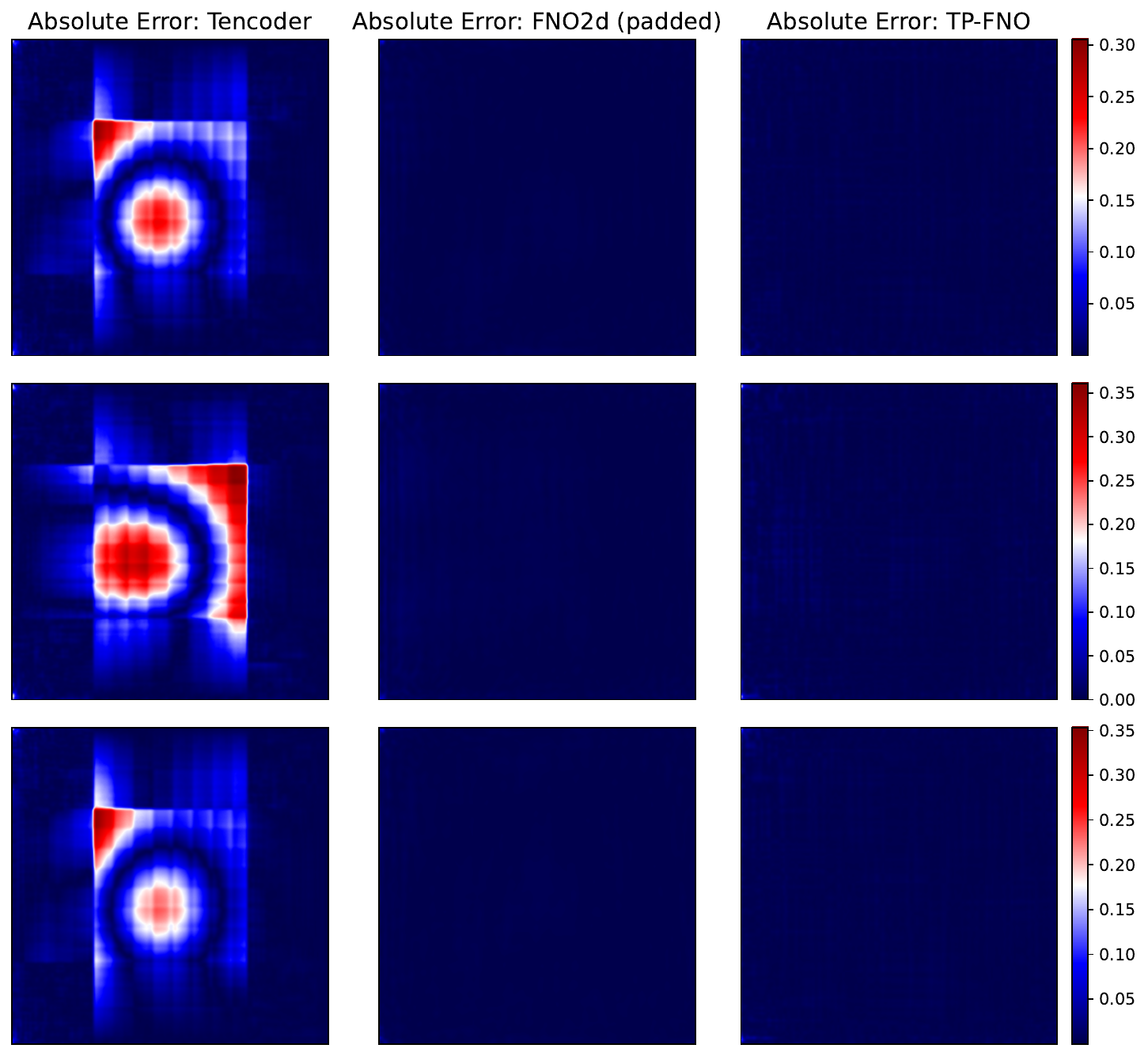}
    \caption{Comparison of the absolute error on in-distribution examples for models trained on 128x128 resolution.}
    \label{fig:mae_in_dist_128}
\end{figure}

Additional visualizations of the model performances on 128x128 can be seen in Figures \ref{fig:field_in_dist_128} and \ref{fig:mae_in_dist_128}.

\subsection{Visualization of Other Out-of-Distribution Results}

\begin{figure} [ht]
    \includegraphics[width=1.05\linewidth]{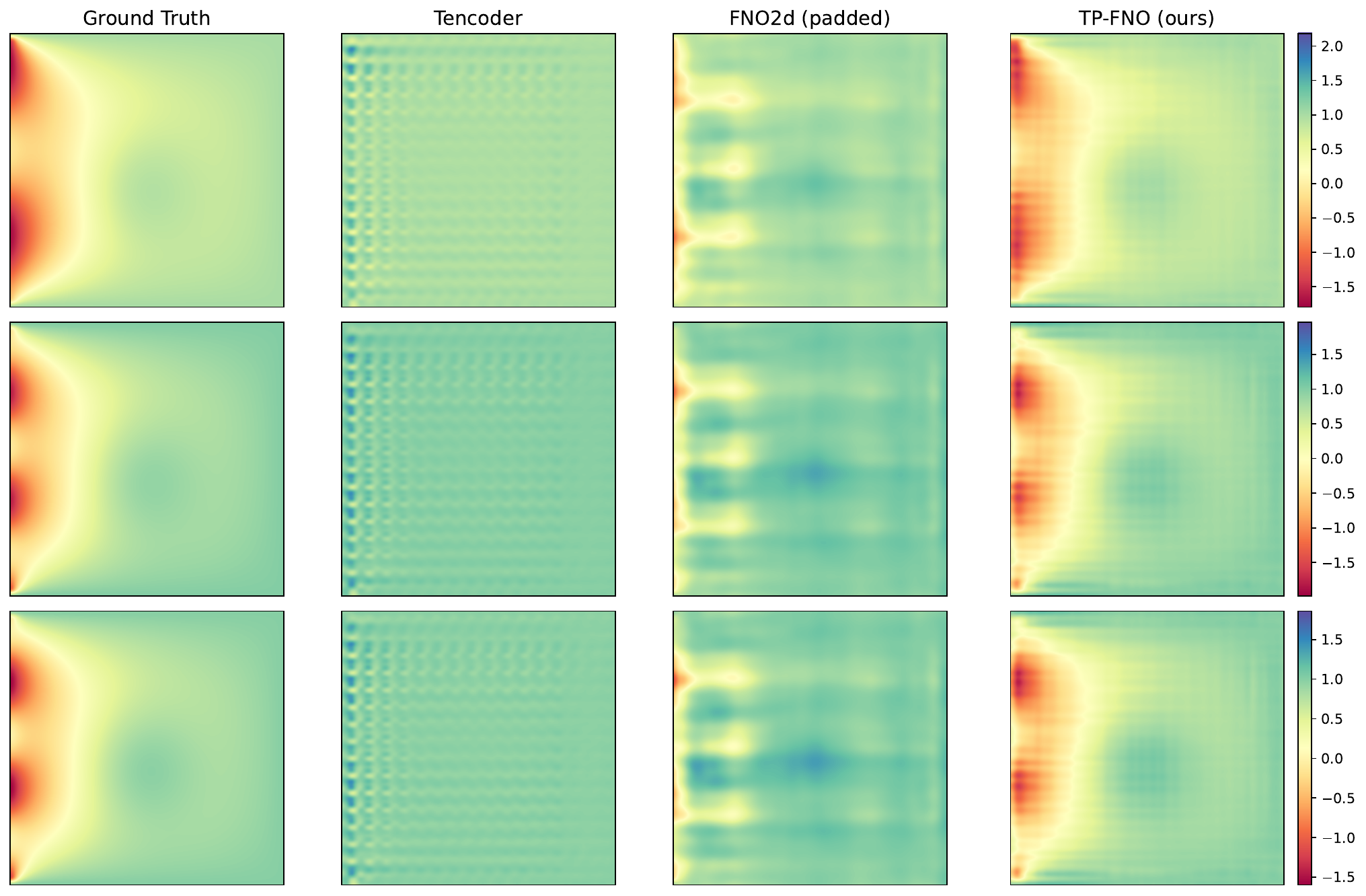}
    \caption{Comparison of the predicted solutions on out-of-distribution sinusoidal examples for models trained on 32x32 resolution and evaluated on 64x64.}
    \label{fig:field_ood_dist_32_fourier}
\end{figure}

\begin{figure}[ht]
    \includegraphics[width=0.95\linewidth]{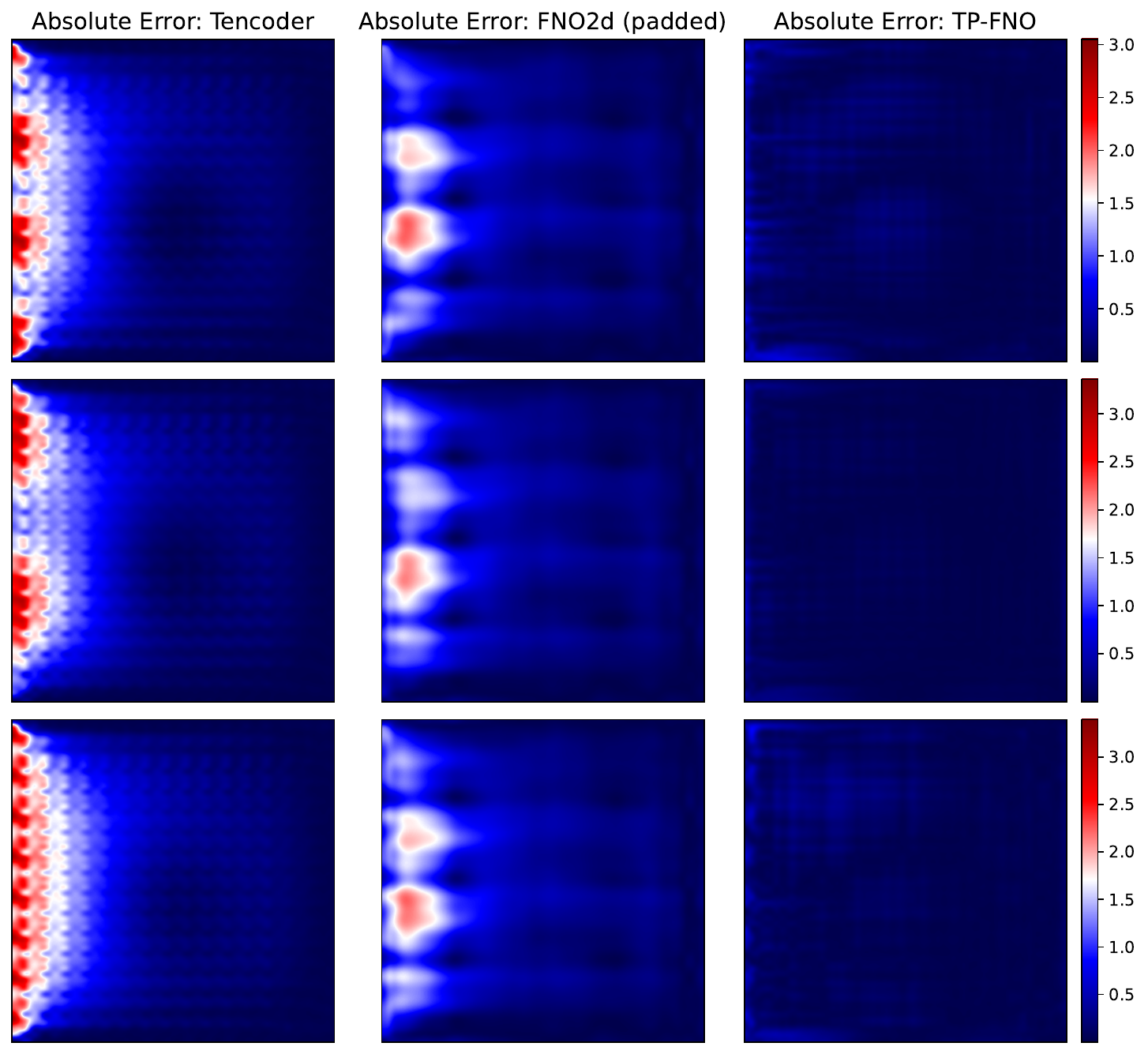}
    \caption{Comparison of the absolute error on out-of-distribution sinusoidal examples for models trained on 32x32 resolution and evaluated on 64x64.}
    \label{fig:mae_ood_dist_32_fourier}
\end{figure}

The out-of-distribution for the sinusoidal examples are shown in \ref{fig:field_ood_dist_32_fourier} and \ref{fig:mae_ood_dist_32_fourier}.

\begin{figure} [ht]
    \includegraphics[width=1.05\linewidth]{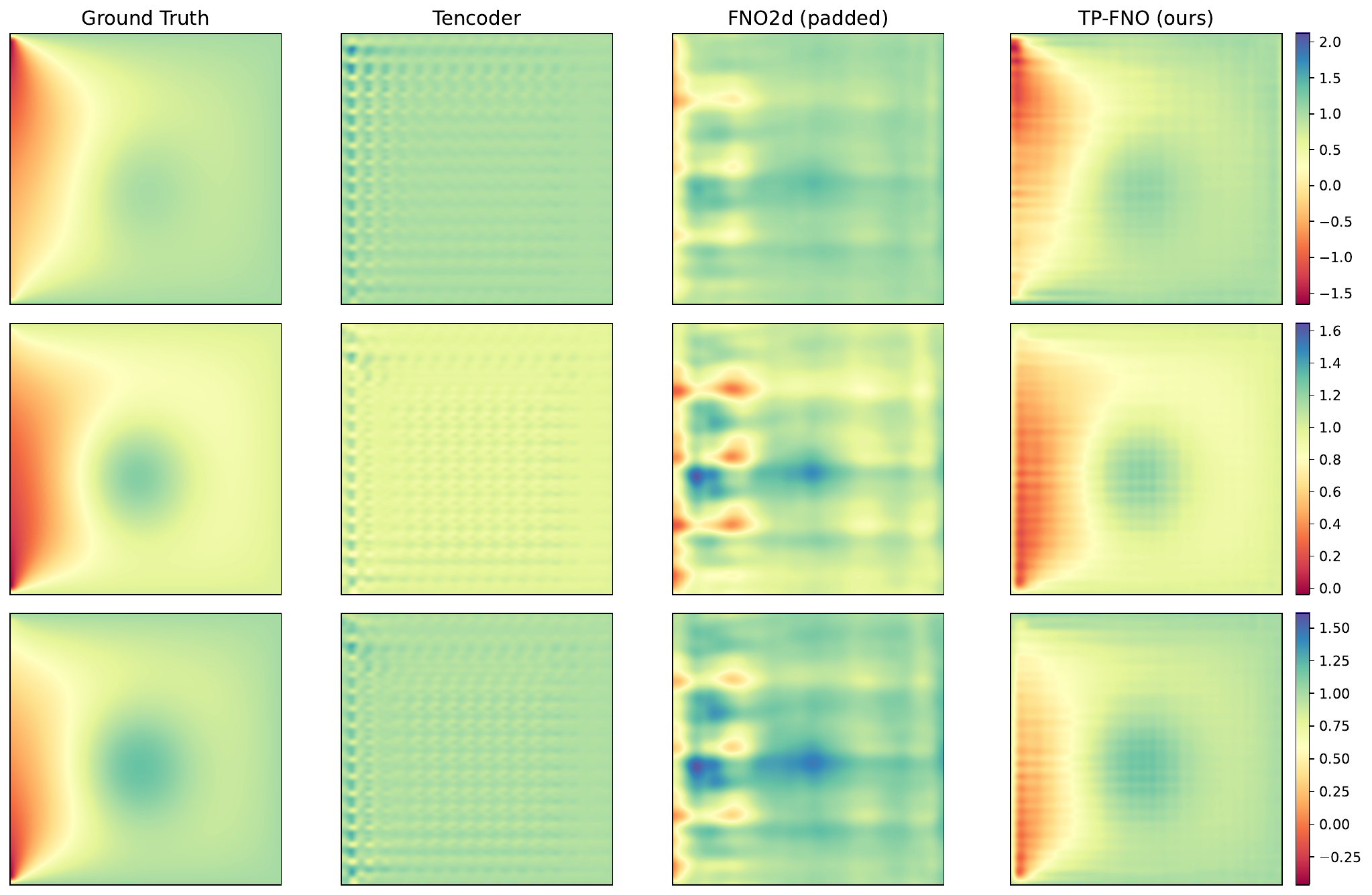}
    \caption{Comparison of the predicted solutions on out-of-distribution polynomial examples for models trained on 32x32 resolution and evaluated on 64x64.}
    \label{fig:field_ood_dist_32_poly}
\end{figure}

\begin{figure}[ht]
    \includegraphics[width=0.95\linewidth]{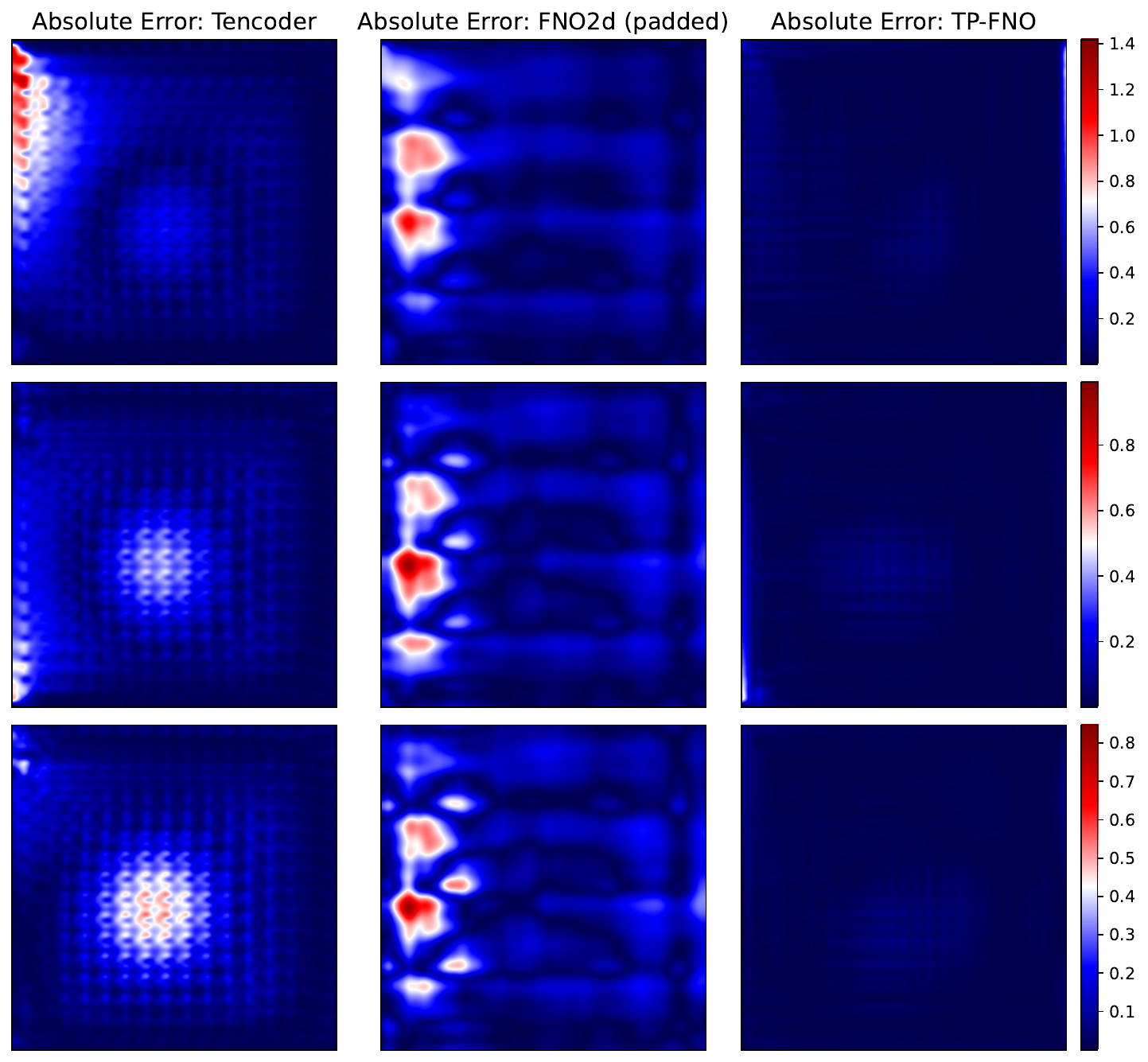}
    \caption{Comparison of the absolute error on out-of-distribution polynomial examples for models trained on 32x32 resolution and evaluated on 64x64.}
    \label{fig:mae_ood_dist_32_poly}
\end{figure}

The out-of-distribution for the polynomial examples are shown in \ref{fig:field_ood_dist_32_poly} and \ref{fig:mae_ood_dist_32_poly}.

\subsection{Convergence of Different Models}
\begin{figure}[ht]
    \includegraphics[width=0.8\linewidth]{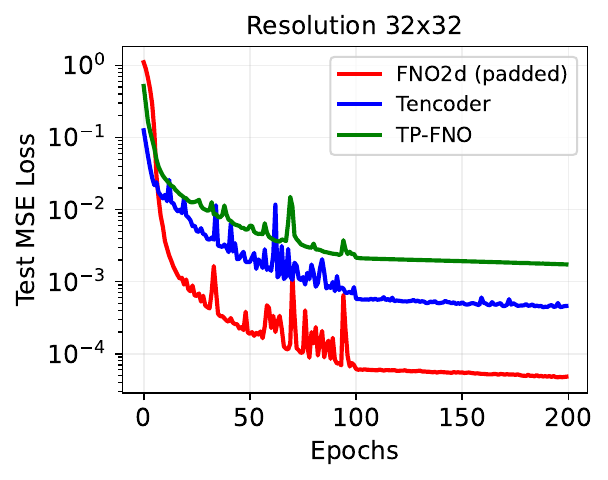}
    \caption{Test MSE loss during training on data of resolution 32x32}
    \label{fig:convergence32}
\end{figure}

\begin{figure}[ht]
    \includegraphics[width=0.8\linewidth]{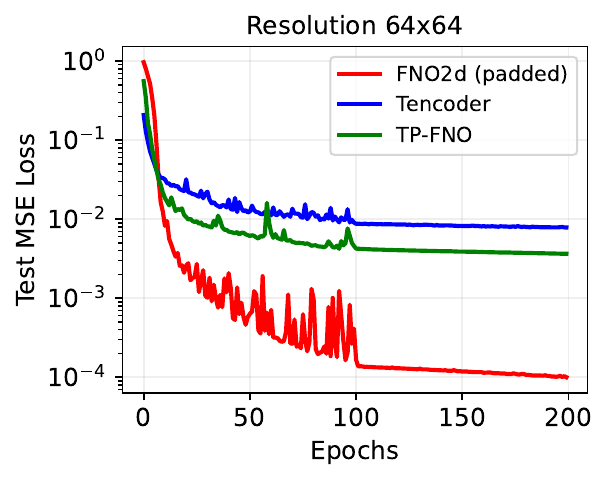}
    \caption{Test MSE loss on data of resolution 64x64}
    \label{fig:convergence64}
\end{figure}

\begin{figure}[ht]
    \includegraphics[width=0.8\linewidth]{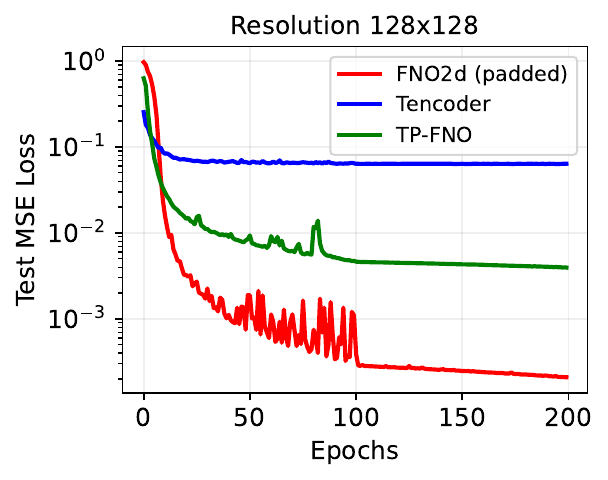}
    \caption{Test MSE loss on data of resolution 128x128}
    \label{fig:convergence128}
\end{figure}

Figure \ref{fig:convergence32}, \ref{fig:convergence64}, \ref{fig:convergence128} shows the convergence of the mean squared error (MSE) loss of the different models while training on the 32x32, 64x64 and 128x128 datasets respectively.

\section{Hyperparameter Details}
For all the models - Tencoder, FNO2d (padded), and our proposed TP-FNO—we used an Adam optimizer with a learning rate of 1e-3, and a Step Learning Rate scheduler with a step size of 100 and $\gamma=0.1$. All models were trained for 200 epochs. Tencoder was trained with a batch size of 8, as recommended in the original paper, while FNO2d and TP-FNO were trained with batch sizes of 128.

Regarding the model architectures, for Tencoder, we adhered to the original architecture, which includes 3 1D CNN layers in the encoder, two 1D CNN layers in the tensor product operation and 5 convolutional transpose layers in the decoder. The Tencoder model had 5,573 parameters. For FNO2d with zero padding, we set the number of Fourier modes to 16 and the width of each Fourier layer to 16, resulting in a model with 527,713 parameters. For our proposed TP-FNO, we set the maximum Fourier modes to 16 and the embedding dimension to 64. The TP-FNO model had 568,241 parameters, making it comparable in size to the FNO2d model.







\end{document}